# QUALITY ESTIMATION OF MACHINE TRANSLATION OUTPUTS THROUGH STEMMING

Pooja Gupta [1], Nisheeth Joshi [2] and Iti Mathur [3]

[1, 2, 3] Department of Computer Science, Apaji Institute, Banasthali University, Rajasthan, India

## ABSTRACT

*Machine Translation is the challenging problem for Indian languages. Every day we can see some machine translators being developed , but getting a high quality automatic translation is still a very distant dream . The correct translated sentence for Hindi language is rarely found. In this paper, we are emphasizing on English-Hindi language pair, so in order to preserve the correct MT output we present a ranking system, which employs some machine learning techniques and morphological features. In ranking no human intervention is required. We have also validated our results by comparing it with human ranking.*

## KEYWORDS

*Machine Translation, Stemming, Machine learning, Language Model.*

## 1. INTRODUCTION

Machine translation is a field which is an amalgamation of areas such as Computational Linguistics, Artificial Intelligence, Translation Theory and Statistics. Machine translation is fast and is available on a click of a button whereas human translation is very slow, time consuming and expensive task as compared to machine translation. But acceptance of machine translation is very low because of reasons like bad translations in available systems and ambiguities. Human languages are highly ambiguous, and produce different meanings in different languages. To overcome this problem we come up with a solution of integrating multiple machine translation engine into one i.e. we create a multi-engine machine translation system. Sometimes it also gives bad results while selecting a final output. So, we need to rank the MT engine outputs for that Manual ranking is acting as a human translation, it's very tedious task.  So we need to perform automatic ranking for a large amount of data with minimum time. In order to develop an automatic ranking system we need to develop several different modules. The very first module of ranking system that comes in machine translation pipeline is N-gram language model. This acts as a baseline system and second module is morphological analysis. In this Stemming/lemmatization are performed. In N-gram LM ranking we used the trigrams approximation approach, Gupta et al. defines this approach [1]. In Stemming based ranking, we used a Hindi Rule based Stemmer. This Stemmer is a simplest morphological parsing system which contains some morphological information. Morphological information is an important part when we consider the design of any MT engine, any natural language processing application or any information retrieval system.

The rest of the paper is organized as follows: In Section2, we briefly give an overview of related work that has been done in this area. Section 3 shows working of stemming done for Hindi language. Section 4 describes our proposed work. In this section we also define corpus creation, algorithms and methodology of ranking approach. Section 5 shows the evaluation and the results of the research. Finally Section 6 gives the conclusion of the paper.

## 2. RELATED WORK

Quality estimation is an open source framework that allows checks several quality indicators. Specia et al. [2] defines the extraction of these indicators from source segments, their translations, external resources like corpora and language models as well as some language tools like parser and part-of-speech tags. Soricut and Narsal [3] used machine learning for ranking the candidate translations; they selected the highest-ranked translation as the final output. Avramidis [4] showed an approach of ranking the outputs using grammatical features. They used statistical parser to analyze and generate ranks for several MT output. Gupta el al. [5] [6] applied a naïve bayes classifier on English-Hindi Machine Translation System and ranked the systems. For evaluating the quality of the system the authors have used some linguistic features. The authors have also compared the results with automatic evaluation metrics. Moore and Quirk [7] described smoothing method for N-gram language models based on ordinary counts for generation language models which can use for quality estimation task. Bharti et al. [8] proposed the work on natural language processing where they gave a detailed study of morphology using paradigm approach.

Stemming was firstly introduced by Julie Beth Lovins [9] in 1968, who proposed the use of it in Natural Language Processing applications. Martin Porter [10] in 1980 improved this stemmer. He suggested a suffix stripping algorithm which is still considered to be a standard stemming algorithm. Goldsmith [11] proposed an unsupervised approach to model morphological variants of European languages. Ramanathan and Rao [12] used the same approach, but used some more rules for stemming for Hindi language. Ameta et al. [13] proposed A Lightweight Stemmer for Gujarati, they showed an implementation of a rule based stemmer of Gujarati and created rules for stemming and the richness in morphology. They further used this stemmer in Gujarati-Hindi machine translation system [14]. Pal et al. [15][16] developed a Hindi lemmatizer which generates rules for removing the affixes along with the addition of rules for creating a proper root word. Gupta et al. [17] Developed a rule based Urdu stemmer which showed an accuracy of 84%. They used this stemmer in evaluating English-Urdu machine translation systems [18][19].

## 3. STEMMING FOR HINDI

Hindi is an Indo – Aryan language and is the official language of India. It is widely spoken by a large number of people of the country. Stemming is the process of reducing a derived word into its stem word or root word by clipping off the unnecessary morphemes. These morphemes are known as suffixes. This suffix stripping is made by generating various rules. This is done by our Hindi rule based stemmer. Our approach learns suffixes automatically from a large vocabulary or dictionary of words extracted from raw text. This vocabulary is known as an exhaustive lexicon list; which contains only root words and derivational words. The purpose of stemming is to obtain the stem of those words which are not found in vocabulary. If stemmed word is present in vocabulary, then that is an actual word, otherwise it may be a proper name or some invalid word. Stemming is used in Information Retrieval systems where input words do not match vocabulary. For example, when a user enters an input word सफलता and if the input word is not present in the vocabulary of the database then it may cause erroneous result. With the help of a stemmer, one can reduce the desired word into its root or stem word. In this example सफल is the stem or root word and ता is the suffix. Stem supplies the main meaning of the word while the suffixes add additional meanings.

## 4. PROPOSED WORK

Our proposed approach is based on n-gram language models. N-gram language models use the Markov assumption to break the probability of a sentence into the product of the probability of each word, given the history of preceding words. We have used Markov chains of order 2 which International Journal on Computational Sciences & Applications (IJCSA) Vol.4, No.3, June 2014are called as trigram approximations. N-gram language models are based on statistics of how likely words are to follow each other. Equations 1, 2 and 3 show the generation of unigram, bigrams and trigrams respectively.

$$P(w_n) = \frac{Count(w_n)}{|V|} \qquad (1)$$

$$P(w_{n-1}w_n) = \frac{Count(w_{n-1}w_n)}{Count(w_{n-1})} \qquad (2)$$

$$P(w_{n-2}w_{n-1}w_n) = \frac{Count(w_{n-2}w_{n-1}w_n)}{Count(w_{n-2}w_{n-1})} \qquad (3)$$

### 4.1. Corpus Creation and Experimental Setup

The approach for creation of corpus is based on language modelling. In language modelling, we have computed the probability of a string then firstly we have been collecting a large amount of text and obtained trigrams along with their number of occurrences or frequency. We have created our ranking system mainly for raw text of tourism domain. However, the corpus also includes words from dictionaries available. It is actually our Bilingual parallel corpus. We used a total of 35000 Hindi sentences giving a total of 513910 unigrams, 308706 bigram word units, and 53062 trigram word units. Another corpus that we have created that is stem corpus of 35000 Hindi sentences.Table1 shows Stemmed trigram corpus of an English sentence and its Hindi translation.

**English Sentence**: Indians must take protective actions to protect their freedom

**Hindi Sentence:** भारताया का अपना सवतत्रता का रक्षा क ालय रक्षात्मक कदम उठान चााहए |

Table1: Stemmed Corpus

| S.No. | Hindi Trigrams | Stem Trigrams |
|---|---|---|
| 1 | भारताया का अपना | भारतीय को अपनी |
| 2 | को अपनी सवतत्रता | का अपना सवतत्र |
| 3 | अपना सवतत्रता का | अपना सवतत्र का |
| 4 | सवतत्रता का रक्षा | सवतत्र का रक्षा |
| 5 | का रक्षा क | का रक्षा के |
| 6 | रक्षा क ालय | रक्षा क ालय |
| 7 | क ालय रक्षात्मक | क ालय रक्षा |
| 8 | ालय रक्षात्मक कदम | ालय रक्षा कदम |
| 9 | रक्षात्मक कदम उठाने | रक्षा कदम उठाना |
| 10 | कदम उठाना चााहए | कदम उठाना चााहए |

17

International Journal on Computational Sciences & Applications (IJCSA) Vol.4, No.3, June 2014

We have used the following algorithms to generate the n-grams for our study. We also generated the stems of corresponding n-grams. We applied these algorithms on both English as well as Hindi sentences separately. These algorithms are shown in Table 2 and Table 3.

**Input:** Raw sentences
**Output**: Annotated Text (N-grams text)

Table2: LM Algorithm

| |
|---|
| Step1.  Input raw sentence file and repeat steps 2 to 4 for each sentence.
Step2.  Split each word of the sentence.
Step3.  Generate trigrams, bigrams and unigrams for the entire sentence.
Step4.  If n-gram is already present than increase the frequency count.
Step5.  If n-gram is unique than it will sort in descending order by their frequencies.
Step6.  Generate Probability of unigrams using equation 1.
Step7.  Generate Probability of bigrams using equation 2.
Step8.  Generate Probability of trigrams using equation 3.
Step9.  Output obtained in file is in our desired n-gram format. |

**Input:** N-grams text
**Output**: Stems text

Table3: Stemming Algorithm

| |
|---|
| Step1.  Input the n-gram word.
Step2.  Matching the word in database.
Step3.  If the word exists in the database then it is displayed as output.
Step4.  If word doesn't exist in the database then the rules are accessed or stripping out the suffix.
Step5.  Rules work by deleting the suffix from the input.
Step6.  Obtained Output in our desired stem format is shown in table 1. |

In our study we have used 1320 English sentences and used six MT engines which were used by Joshi [20] in his study. The list of engines is shown in table 4. Among these E1, E2 and E3 are MT engines freely available on the internet. E4, E5 and E6 are MT engines that we developed using different MT toolkits. E4 was a MT system which was trained using Moses MT toolkit [21]. This system used syntax based model [22]. We used Collins parser to generate parses of English sentences and used a tree to string model to train the system. E5 was a simple phrase based MT system which also used Moses MT toolkit. E6 was an example based MT system that was developed by Joshi et al. [23] [24]. These three systems used the 35000 English-Hindi parallel corpora to train and tune themselves. We used 80-20 ratio for training and tuning i.e. we used 28000 sentences to train the systems and remaining 7000 sentences to tune the systems.

18International Journal on Computational Sciences & Applications (IJCSA) Vol.4, No.3, June 2014

We have used the following algorithms to generate the n-grams for our study. We also generated the stems of corresponding n-grams. We applied these algorithms on both English as well as Hindi sentences separately. These algorithms are shown in Table 2 and Table 3.

**Input:** Raw sentences
**Output**: Annotated Text (N-grams text)

Table2: LM Algorithm

- Step1.  Input raw sentence file and repeat steps 2 to 4 for each sentence.
- Step2.  Split each word of the sentence.
- Step3.  Generate trigrams, bigrams and unigrams for the entire sentence.
- Step4.  If n-gram is already present than increase the frequency count.
- Step5.  If n-gram is unique than it will sort in descending order by their frequencies.
- Step6.  Generate Probability of unigrams using equation 1.
- Step7.  Generate Probability of bigrams using equation 2.
- Step8.  Generate Probability of trigrams using equation 3.
- Step9.  Output obtained in file is in our desired n-gram format.

**Input:** N-grams text
**Output**: Stems text

Table3: Stemming Algorithm

- Step1.  Input the n-gram word.
- Step2.  Matching the word in database.
- Step3.  If the word exists in the database then it is displayed as output.
- Step4.  If word doesn't exist in the database then the rules are accessed or stripping out the suffix.
- Step5.  Rules work by deleting the suffix from the input.
- Step6.  Obtained Output in our desired stem format is shown in table 1.

In our study we have used 1320 English sentences and used six MT engines which were used by Joshi [20] in his study. The list of engines is shown in table 4. Among these E1, E2 and E3 are MT engines freely available on the internet. E4, E5 and E6 are MT engines that we developed using different MT toolkits. E4 was a MT system which was trained using Moses MT toolkit [21]. This system used syntax based model [22]. We used Collins parser to generate parses of English sentences and used a tree to string model to train the system. E5 was a simple phrase based MT system which also used Moses MT toolkit. E6 was an example based MT system that was developed by Joshi et al. [23] [24]. These three systems used the 35000 English-Hindi parallel corpora to train and tune themselves. We used 80-20 ratio for training and tuning i.e. we used 28000 sentences to train the systems and remaining 7000 sentences to tune the systems.

18

Table 4. MT Engines

| Engine No. | Description |
|---|---|
| E1 | Microsoft Bing MT Engine[1] |
| E2 | Google MT Engine[2] |
| E3 | Babylon MT Engine[3] |
| E4 | Moses Syntax Based Model |
| E5 | Moses Phrase Model |
| E6 | Example Based MT Engine |

**4.2. Methodology**

To rank MT outputs of the various systems we first generated the trigrams of English sentence as well as its translations produced by different MT engines. After that we applied stemming algorithm and got stemmed sentences. Then we generated the stem trigrams of all translations. To rank the translations we applied the following algorithm:

**Input:** English Sentence with MT outputs

**Output:** Ranked MT output list

**Ranking Algorithm**

Step1. Trigrams from English sentences are generated.
Step2. These trigrams are matched with English language model and matched ones are retained.
Step3. Match retained English trigram's lexicons with English-Hindi parallel lexicon list and it match with Hindi stem trigram's lexicon list.
Step4. If a match is found then register corresponding Hindi stem trigram lexicon.
Step5. Match Hindi language model with registered Hindi stem lexicons and sum the probabilities of each match.
Step6. Perform these steps on all MT outputs.
Step7. Sort MT outputs in descending order with respect to their cumulative probabilities.

To have a better understanding of the functionality, we have illustrated the entire process through the following example.

**Sentence:** The Indian Himalayan range is undoubtedly one of the most spectacular and impressive mountain ranges in the world.

**E1 Output:** भारतीय हिमालय रज निस्सदेह दुनया म सबसे शानदार और प्रभावशाला पवत श्रृंखला म से एक है।

**E2 Output:** भारतीय हिमालय पवत श्रृंखला बेशक दुनया म सबसे शानदार और प्रभावशाला पवत श्रृंखला म से एक है.

---

[1] http://www.microsofttranslator.com
[2] http://translate.goolge.com
[3] http://translation.babylon.com

International Journal on Computational Sciences & Applications (IJCSA) Vol.4, No.3, June 2014

**E3 Output:** भारतीय हिमालय पर्वत का यह निस्सदेह रूप से एक सबसे भव्य एवं प्रभावशाला पर्वत शृंखलाओं विश्व म ।

**E4 Output:** यहाँ भारतीय हिमालया शृखला है undoubtedly एक के के अधिकाश पहाड़ी और impressive पर्वत शृखलाए म के world.

**E5 Output:** The Indian Himalayan शृखला का एक undoubtedly है के सबसे spectacular और impressive mountain ranges विश्व के म है ।

**E6 Output:** भारतीय हिमालया शृखला है निःसन्देह समूचे एक का सबसे देखते हा बनती और पर्वत शृखलाआ म विश्व ।

Table 5 shows the n-gram statistics of these sentences and also shows the sum of cumulative probabilities of these trigrams. By looking at the data we can rank the system according to their probabilities.

Table 5. MT Systems

| Engine | Unigrams | Bigrams | Trigrams | Prob. Sum |
|--------|----------|---------|----------|-----------|
| E1 | 16 | 15 | 14 | 0.843723 |
| E2 | 17 | 16 | 15 | 0.843723 |
| E3 | 17 | 16 | 15 | 0.574318 |
| E4 | 18 | 17 | 16 | 0.0 |
| E5 | 21 | 20 | 19 | 0.293709 |
| E6 | 18 | 17 | 16 | 0.463309 |

## 5. EVALUATION

To evaluate the performance of our system we collected 1300 sentences from tourism domain. These sentences were not part of 35000 sentences that were used to train the models. To validate our results we compared the ranks of the system with the ranks given to MT systems by a human evaluator. The human evaluator used a subjective human evaluation metric that was developed by Joshi et al. [25]. This metric evaluated an MT output on eleven parameters. These were:

1. Translation of Gender and Number of the Noun(s).
2. Identification of the Proper Noun(s).
3. Use of Adjectives and Adverbs corresponding to the Nouns and Verbs.
4. Selection of proper words/synonyms (Lexical Choice).
5. Sequence of phrases and clauses in the translation.
6. Use of Punctuation Marks in the translation
7. Translation of tense in the sentence
8. Translation of Voice in the sentence
9. Maintaining the semantics of the source sentence in the translation
10. Fluency of translated text and translator's proficiency
11. Overall quality of the translation

Each MT outputs were adjudged on these 11 parameters. The human evaluator was asked to give a score on a 5-point scale. The scale is shown is Table 6.

Table 6. Human Evaluation Scale

| Score | Description |
|---|---|
| 1 | Ideal |
| 2 | Perfect |
| 3 | Acceptable |
| 4 | Partially Acceptable |
| 5 | Not Acceptable |

For evaluation, we used the methodology used by Joshi et al.[26]. We evaluated the system generated ranks with human ranks in two different categories. At first we compared the ranks of all the systems, irrespective of their type. In second category we compared the ranks of only web based systems and in third category we compared the ranks of only MT toolkits or system which had very limited corpora to train and tune themselves.

In combined category, engine E1 performed better than any other MT engine. It scored the highest rank. Out of 1300 sentences, it managed to score highest rank for 407 sentences. Engine E2 was the second best while engines E4 did not performed so well. Table 7 shows the results of this study.

Table 7. Ranking at Combined Category

| Engine | STEM LM Ranking | Human Ranking |
|---|---|---|
| **E1** | **407** | **376** |
| E2 | 285 | 279 |
| E3 | 145 | 140 |
| E4 | 8 | 7 |
| E5 | 256 | 205 |
| E6 | 236 | 240 |

In web-based category, again E1 and E2 performed better and were the top ranking systems while E4 was the worst. Table 8 shows the results of this study. In MT Toolkits category, E6 performed better than other MT engines and E4 was the worst engine. Table 9 shows the results of this study. These ranks were similar to the ranks provided by human evaluator. Figure 2, 3 and 4 summarized these data.

Table 8. Ranking at Web-Based Category

| Engine | Stem LM Ranking | Human Ranking |
|---|---|---|
| **E1** | **603** | **587** |
| E2 | 432 | 473 |
| E3 | 235 | 145 |

Table 9. Ranking at MT Toolkits Category

| Engine | Stem LM Ranking | Human Ranking |
|---|---|---|
| E4 | 16 | 18 |
| E5 | 234 | 254 |
| **E6** | **356** | **288** |

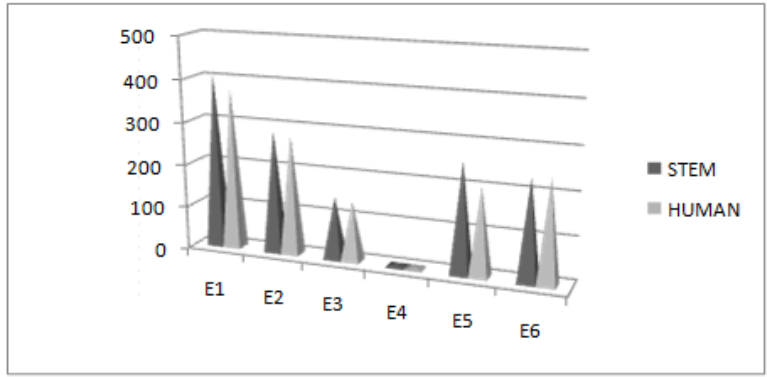

Figure 1. Ranking at Combined Category

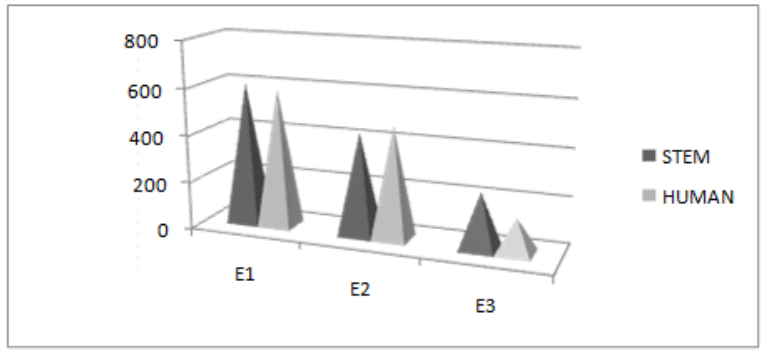

Figure 2. Ranking at Web-Based Category

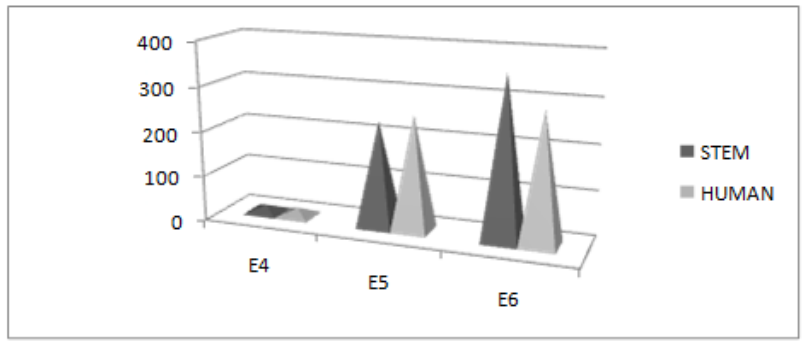

Figure 3. Ranking at MT Toolkits Category

## 6. CONCLUSIONS

In this paper, we have shown the effective use of language models and morphological analysis in ranking MT systems. For this we had generated language models for English, Hindi sentences as well as for Hindi Stemmed Text. The system described here are very simple and efficient for automatic ranking even when the amount of available raw text is not so large. It was found that the ranks produced by stem language model based ranking and the ranks of human judge were similar. We found that Microsoft Bing translator was best translator among six of them as it gave correct translated sentence. The best performance of the current system is as good as the baseline system. The stemmed language model based ranking have a much higher accuracy than the baseline model. The time taken by stem based ranking is maximum as compared to baseline and human ranking. Moreover as an immediate future study we can incorporate parts of speech tagging into language models and then perform the ranking and see if the performance of the system improves or not.

**AUTHORS**

Pooja Gupta is pursuing her M.Tech in Computer Science from Banasthali University, Rajasthan and is working as a Research Assistant in English-Indian Languages Machine Translation System Project sponsored by TDIL Programme, DeitY. She has her interest in Machine Translation specifically in English-Hindi Language Pair. Her current research interest includes Natural Language Processing and Machine Translation.

Dr. Nisheeth Joshi is an Associate Professor at Banasthali University. He has been primarily working in design and development of evaluation Matrices in Indian languages. Besides this he is also actively involved in the development of MT engines for English to Indian Languages. He is one of the experts empanelled with TDIL programme, Department of Electronics and Information Technology (DeitY), Govt. of India, a premier organization which foresees Language Technology Funding and Research in India. He has several publications in various journals and conferences and also serves on the Programme Committees and Editorial Boards of several conferences and journals.

Iti Mathur is an Assistant Professor at Banasthali University. Her primary area of research is Computational Semantics and Ontological Engineering. Besides this she is also involved in the development of MT engines for English to Indian Languages. She is one of the experts empanelled with TDIL Programme, Department of Electronics and Information Technology (DeitY), Govt. of India, a premier organization which foresees Language Technology Funding and Research in India. She has several publications in various journals and conferences and also serves on the Programme Committees and Editorial Boards of several conferences and journals.